\documentclass[10pt,twocolumn,letterpaper]{article}

\usepackage{cvpr}      

\usepackage{graphicx}
\usepackage{amsmath}
\usepackage{amssymb}
\usepackage{amsthm}
\usepackage{booktabs}
\usepackage{multicol}
\usepackage{multirow}
\usepackage[accsupp]{axessibility}  

\newtheorem{prop}{Proposition}
\newtheorem{define}{Definition}

\newcommand{\mb}[1]{\mathbf{#1}}
\newcommand{\mc}[1]{\mathcal{#1}}

\usepackage[pagebackref,breaklinks,colorlinks]{hyperref}

\usepackage[capitalize]{cleveref}
\crefname{section}{Sec.}{Secs.}
\Crefname{section}{Section}{Sections}
\Crefname{table}{Table}{Tables}
\crefname{table}{Tab.}{Tabs.}

\def\cvprPaperID{6373} 
\def\confName{CVPR}
\def\confYear{2022}

\begin{document}

\title{Learning Distinctive Margin toward Active Domain Adaptation}

\author{Ming Xie\textsuperscript{1}\footnotemark[1]\quad Yuxi Li\textsuperscript{2}\footnotemark[1]\quad Yabiao Wang\textsuperscript{2}\quad Zekun Luo\textsuperscript{2}\quad Zhenye Gan\textsuperscript{2}\\
Zhongyi Sun\textsuperscript{2}\quad Mingmin Chi\textsuperscript{1}\footnotemark[2]\quad Chengjie Wang\textsuperscript{2}\footnotemark[2]\quad Pei Wang\textsuperscript{3}\\
\textsuperscript{1}Fudan University \quad \textsuperscript{2}Tencent Youtu Lab \quad \textsuperscript{3}NAOC CAS\\
{\tt \small{\{mxie20,mmchi\}@fudan.edu.cn}}\\
{\tt \small{\{yukiyxli,caseywang,zekunluo,wingzygan,zhongyisun,jasoncjwang\}@tencent.com}}\\
{\tt \small{wangpei@nao.cas.cn}}
}
\maketitle
\renewcommand{\thefootnote}{\fnsymbol{footnote}}
\footnotetext[1]{Both author contributed equally to this work. Work is completed during Ming Xie's internship at Tencent Youtu Lab.}
\footnotetext[2]{Corresponding author}

\begin{abstract}
    Despite plenty of efforts focusing on improving the domain adaptation ability (DA) under unsupervised or few-shot semi-supervised settings, recently the solution of active learning started to attract more attention due to its suitability in transferring model in a more practical way with limited annotation resource on target data. Nevertheless, most active learning methods are not inherently designed to handle domain gap between data distribution, on the other hand, some active domain adaptation methods (ADA) usually requires complicated query functions, which is vulnerable to overfitting. In this work, we propose a concise but effective ADA method called Select-by-Distinctive-Margin (SDM), which consists of a maximum margin loss and a margin sampling algorithm for data selection. We provide theoretical analysis to show that SDM works like a Support Vector Machine, storing hard examples around decision boundaries and exploiting them to find informative and transferable data. In addition, we propose two variants of our method, one is designed to adaptively adjust the gradient from margin loss, the other boosts the selectivity of margin sampling by taking the gradient direction into account. We benchmark SDM with standard active learning setting, demonstrating our algorithm achieves competitive results with good data scalability. Code is available at \textcolor{blue}{\url{https://github.com/TencentYoutuResearch/ActiveLearning-SDM}}
\end{abstract}

\section{Introduction}

\begin{figure}
    \centering
    \includegraphics[width=0.48\textwidth]{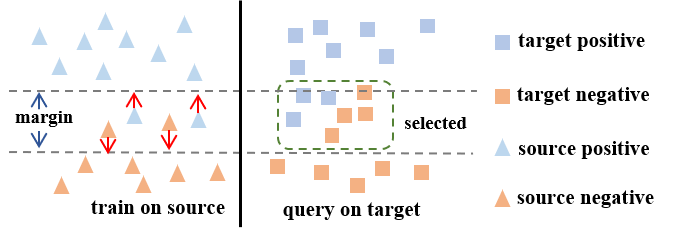}
    \caption{A simple conceptual illustration of our Select-by-Distinctive-Margin pipeline. Before each sampling step, a model is trained with a maximum margin objective, and unlabeled data lying in the margin with similar distance to different categorical centers are sampled to augment training data.}
    \label{fig:concept}
    \vspace{-3mm}
\end{figure}

The domain adaptation problem has been widely studied in transfer learning society, where adaptation algorithms are designed to generalize a model trained on source domain to a target domain with different data distribution~\cite{ben2010theory}. In most of studies, the semantic labels from target domain are assumed to be unavailable~\cite{ben2010theory,dann,cdan,bnm,ent} (UDA) or only few-shot of target samples are labeled~\cite{mme,atdoc,li2021learning}(SSDA). However, in a more practical sense, although it is difficult to annotate all data in target domain, a moderate amount of labeled data should be acceptable given certain budget on annotations cost. 

With this consideration, domain adaptation turns into an active learning problem (AL), which focusing on additionally labeling limited data to bring maximum improvement of machine learning algorithms~\cite{sener2018active,sinha2019variational,kim2021task,yoo2019learning,Zhang_2020_CVPR,choi2021vab,badge}. However, currently most active learning algorithms are derived from a pure semi-supervised scenario, where the unlabeled data are assumed to conform to the same distribution as labeled data. These methods usually focus on designing a distinctive query function to depict how informative or representative an unlabeled data sample is, which highly relies on the uncertainty~\cite{choi2021vab,sinha2019variational} or structural distribution of data features~\cite{sener2018active,badge}. On the contrary, in a domain adaptation problem, the task model is initially trained with only source data and the query function is usually correlated to the prediction of task models, in this case, most of target data will be discriminated as uncertain regardless its location in feature space. Consequently, the sampling methods are prone to sample some target samples that are easily classified and make less effect on the biased decision boundaries.

Recently, there exist some researches aimed at appropriate data selection under the scenario of domain adaptation. However, these methods either design complicated and hand-crafted query function with deliberately designed architecture~\cite{fu2021transferable,S3VAADA}, or select data in a tedious manner of high complexity~\cite{CLUE}. These complicated design makes the query function and selection strategies easy to overfit to a certain transferring scenario and hard to be extended to more general cases. In addition, most of these methods simply exploit all source data equally during training~\cite{su2020active,fu2021transferable,CLUE}, which is vulnerable to bias toward source domain and results unreliable query. Besides, few of studies above discuss the intrinsic relationship between their training objective and query function, ignoring the potential correlation between data of two domains during selection.

With the consideration above, in this paper, we propose to tackle domain adaptation problem with a simple but effective  active learning strategy called Selective-by-Distinctive-Margin (SDM) by evaluating the distance from a data sample to different categorical clusters (as shown in Figure~\ref{fig:concept}). Different from most of previous efforts focusing on selecting data through the uncertainty or diversity of pure unlabeled target data~\cite{sener2018active,sinha2019variational,fu2021transferable,CLUE,S3VAADA}, SDM makes attempt to select unlabeled data via their relation to some ``hard examples'' from the source domain. However, instead of explicitly model such data relation, we implicitly depict the similarity between unlabeled samples and potential hard source samples via a simple maximum margin loss function. \emph{Intuitively}, the margin loss will guide the network to maximize the distance between close examples from different categorical clusters in source domain, meanwhile ignore the affect from well classified source samples. This reversely helps detecting informative target samples still lying near the trained decision boundaries through a simple margin sampling query function. By collecting these data into training set, the manifold of decision boundaries can be further refined and generalized to target distribution. \emph{Theoretically}, by analyzing with a simplified linear model, we confirm that model trained with margin loss can act like a Support Vector Machine~\cite{svm}, which collects only ``hard examples'' in source domain, and take these examples to detect unlabeled target data via the similarity in feature space.

In addition, derived from the simple SDM baseline, we further extend the strategy into two variants. For the training phase, for the sake of dynamically adjust gradient of margin loss to adapt to samples of different difficulties, we propose to extend the original margin loss to a dynamic form with adaptive modulation factor and max-logit reglularizer. On the other hand, during sample selection, to boost the selectivity, we take the first-order gradient of margin sampling function as additional guidance in query function, leading to select target samples which decreases the sampling function in the fastest direction with its estimated gradient. Further, both variants can further be combined together to construct more effective active learning pipeline.

Our SDM algorithm is evaluated on different domain adaption benchmarks like Office-Home~\cite{saenko2010adapting} and Office-31~\cite{venkateswara2017deep} under a classical active learning setting, besides, we also extend our method to a general active learning task on CIFAR-10~\cite{krizhevsky2009learning}, demonstrating our approach can achieve state-of-the-art results with less query complexity and good data-scalablility. In a nutshell, our contributions can be summarized into three folds:
\begin{itemize}
    \item We propose Select-by-Distinctive-Margin (SDM), a concise but effective active learning method for active domain adaptation, which consists of a maximum margin loss and a margin sampling function as a complete active learning cycle. Theoretical analysis is provided to show this SDM framework work like a SVM to take hard examples to mine informative targets.
    \item Derived from the SDM baseline, two variants are developed. One is designed for training phase to dynamically adjust margin loss gradient, the other is designed to enhance the selectivity with the help of first-order gradient of margin sampling function. 
    \item Experiments conducted on several domain adaptation benchmarks show that our approach can achieve state-of-the-art results with limited annotation budget. 
\end{itemize}

\section{Related Work}

\noindent\textbf{Domain Adaptation.} The goal of domain adaptation is to generalize a model trained on source domain to target data distribution~\cite{ben2010theory}. The core issue of domain adaptation lies in the misalignment between feature and label space of source and target domain. To deal with this problem, previous domain adaptation focus on guiding a deep neural network to learn some domain invariant representation and classifiers. To be specific, the adversarial training~\cite{dann,cdan} is utilized to align feature distribution with a domain discriminator, regularizers like entropy constraint~\cite{ent,mme} or maximum prediction rank~\cite{bnm} are applied to implicitly constrain the cross-domain feature space. Recently, there are also some works regard the domain alignment as minimizing the one-to-one optimal matching cost across two sets~\cite{damodaran2018deepjdot}.

One common characteristic of methods above is that all of them assume the annotation in target domain is not accessible or only accessible for a few data, resulting unsupervised or semi-supervised domain adaptation setting. However, in a more practice scene, a moderate number of labeled data from target domain is usually allowed, and there are already some pseudo-label based methods demonstrating some properly labeled target data are powerfully enough to adapt a model from source to target domain~\cite{mixmatch,sohn2020fixmatch,atdoc,Li_2021_ICCV}. As a result, new demand emerges to maximize the model transfer ability given a proper budget of annotated target data samples, which is highly overlapped with the study interests of Active Learning community.

\noindent\textbf{Active Learning.} The research of active learning aims at selecting proper samples to label and taking them to augment original training set and maixmum the improvement on model performance~\cite{settles1995active}. To measure the value of labeling a sample, a query function is usually designed to assign a query score to each sample for rank and selection. Classically, the query function is decided by the uncertainty metrics like entropy, score margin~\cite{balcan2007margin} or least confidence~\cite{lewis1994heterogeneous}. Recently, some advanced active learning pipelines are proposed, which are usually accompanied with deliberately designed training process, among which the Variational Auto Encoder is widely used to model the probability of erroneous prediction~\cite{choi2021vab} or directly learn a binary classifier~\cite{sinha2019variational,Zhang_2020_CVPR} or sample loss ranker~\cite{kim2021task,yoo2019learning} to select samples. Besides, there are other studies starting from the coverage of appended samples, and select data toward the objective of maximum diversity~\cite{sener2018active,badge}. All the methods above achieve promising performance on active learning task of consistent data distribution, however, none of them is designed with specific consideration of potential domain gap between labeled and unlabeled data. Consequently, these query function or sampling strategy are easy to select data with less training difficulty. 

\noindent\textbf{Active Domain Adaptation.} AADA~\cite{su2020active} is one of the earliest research to apply active learning technique specifically for domain adaptation, which applies a discriminator with cross-domain adversarial learning to construct sample query function. The work of~\cite{fu2021transferable,S3VAADA} consider the domain misalignment and design series of training objective and rules to measure the uncertainty and domainness of a target sample,~\cite{fu2021transferable} further proposes a randomize selection strategy to enhance the sample diversity. The method of CLUE~\cite{CLUE} design a entropy weighted clustering algorithm to take both diversity and uncertainty of target data into an unified clustering framework. 

Nevertheless, most of these approaches rely on scenario-specific prior and complicated query functions with series of hyper-parameters, making the methods easy to overfit to specific transfer scenarios and not general. Besides, there are some complicated operations like adversarial example~\cite{S3VAADA,fu2021transferable} or clustering~\cite{CLUE} with high complexity. In contrast, our SDM algorithm is simple in both training and data selection with insightful theoretical interpretation, by exploiting only some hard examples from source domain, our strategy can achieve promising results on different benchmarks.

\section{Approach}
\begin{figure}
    \centering
    \includegraphics[width=0.48\textwidth]{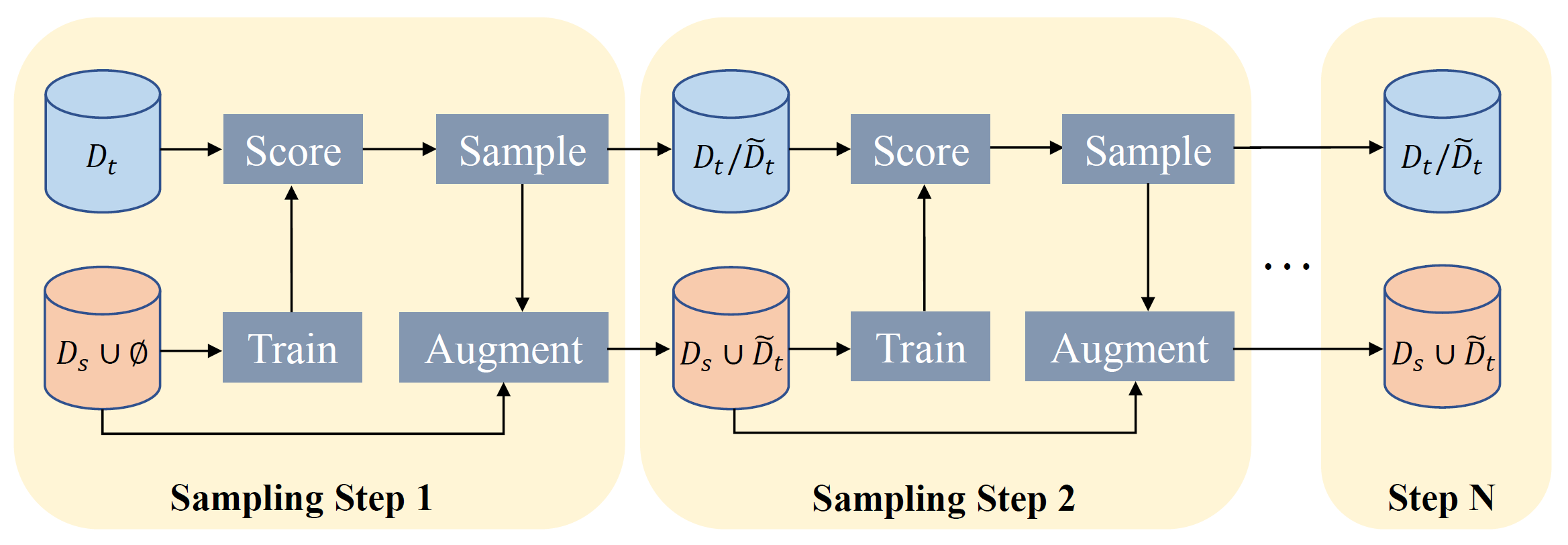}
    \caption{Illustration of active learning loop for domain adaptation}
    \label{fig:train_step}
\end{figure}
\subsection{Problem Formulation}
In the problem of Active Domain Adaptation, a labeled source domain is denoted as $\mc{D}_s=\{(x_s, y_s)\}$, with data $x_s$ and its semantic label $y_s \in \{1,2,\cdots,K\}$, where $K$ is the number of class types, an unlabeled target domain is denoted as $\mc{D}_t=\{x_t\}$. Meanwhile, we denote a labeled target set as $\widetilde{\mc{D}}_t$, which is an empty set $\phi$ initially. With these initial data and a given annotation budget $B$, an active domain adaptation loop can be build as Figure~\ref{fig:train_step}. The unlabeled data is sampled several times, for each selected data $\hat{x}_t \in \mc{D}_t/\widetilde{\mc{D}}_t$, annotators will assign its label $\hat{y}_t$ to it, and $\widetilde{\mc{D}}_t$ is augmented with new labeled target data $\{(\hat{x}_t, \hat{y}_t)\}$ after each sampling step, then the model can be trained with $\mc{D}_s\cup\widetilde{\mc{D}}_t$, afterward the updated model is exploited to select new target data for annotation from set $\mc{D}_t/\widetilde{\mc{D}}_t$. The process repeats until appended number of target samples achieves the budget $|\widetilde{\mc{D}}_t|=B$. For the ease of representation, we denote our model as a composition of a feature extractor $g(\cdot)$ to extract data feature $\mb{f}=g(x)$ and a linear classifier $c(\cdot)$ to categorize a feature into class logit vector of size $K$.

\subsection{Select by Distinctive Margin}
\subsubsection{Pipeline} 
In a classical paradigm, all labeled data from $\mc{D}_s\cup\widetilde{\mc{D}}_t$ can be utilized to train a new deep network, which is widely followed by previous ADA approaches~\cite{fu2021transferable,S3VAADA,CLUE}. However, such strategy makes trained model dominated and biased toward some salient area in source domain of high data density at early stages, and reversely prevent the query function from detecting informative target data.

To mitigate such source-oriented bias, we propose to exploit only ``hard examples'' from source domain to construct our training objective, since these examples are important to shape the decision boundaries with less domain-biased information. Therefore we design the categorical-wise margin loss to supervise the network output due to its inherent selective property
\begin{equation}\label{eq:margin_loss}
    \mc{L}_m(x, y) = \sum_{i\neq y}{\left[m - c(g(x))_y + c(g(x))_i\right]_+}
\end{equation}
where $[x]_+$ denotes zero-clip operation $max(0, x)$, the subscript $y$ and $i$ indicates the $y$-th and $i$-th entry of vectors, and $m$ is a hyperparameter to control the expected margin width. From Eq~(\ref{eq:margin_loss}), we see only samples with similar classification score between ground-truth class and other classes can contribute the gradient for deep network, thus the model will not be dominated by redundant source samples and be easier to transfer to target domain.

On the other hand, since the loss in Eq~(\ref{eq:margin_loss}) explicitly enlarges the gap between different category clusters, it is natural to pay more attention to those samples with smaller gap between category-scores in target domain due to the impact they will have on current learned decision boundaries. As a result, a margin sampling query function is proposed to evaluate the importance of an unlabeled target sample
\begin{align}
    \mb{p} & = \textbf{softmax}(c(g(x_t))) \\
    Q(x) & = 1-\left(\mb{p}_{1^*} - \mb{p}_{2^*}\right) \quad \forall x_t \in \mc{D}_t/\widetilde{\mc{D}}_t\label{eq:query}
\end{align}
where the subscript $1^*$ and  $2^*$ indicates the index of maximum and second maximum value of a vector, before computing the query function, softmax operation is applied to map the logit vector to normalized probability to ensure the scale of $Q(x) \in (0, 1)$. The smaller category gap a sample has, the larger $Q(x)$ value is assigned. Therefore unlabeled target data can be re-ranked by the metric in Eq~(\ref{eq:query}) and top ranked samples are labeled to augment training set.

\subsubsection{Theoretical Insights} 
To further discuss how a marginal loss helps our model to select informative sample under the margin sampling query function, we simplified our model into a parameterized binary linear classification problem $c(g(x))=[w_+,w_-]^Tx$, where a data feature $x \in \mc{R}^D$ can only be categorized as positive or negative, under this setting, we prove that \textbf{the query function $Q(x)$ is correlated to the similarity between $x$ and ``hard examples'' during training}. To be specific, we denote a training batch as $\mc{S}=\mc{S}^+\cup\mc{S}^-$, where $\mc{S}^+$ denotes the set of positive sample and $\mc{S}^-$ contains all negative samples in a batch, these training samples are exploited to train a binary linear classifier with positive weight $w_+$ and negative weight $w_-$ via a margin loss as Eq~(\ref{eq:margin_loss}), after training, a sample $x$ can be discriminated via its predicted probability of belonging to a certain category
\begin{equation}\label{eq:prob}
    p^+(x) = \frac{e^{w_{+}^{T}x}}{e^{w_{+}^{T}x}+e^{w_{-}^{T}x}} \quad p^-(x) = \frac{e^{w_{-}^{T}x}}{e^{w_{+}^{T}x}+e^{w_{-}^{T}x}}
\end{equation}

\noindent\textbf{Data Selection.} With the formulation in Eq~(\ref{eq:prob}), we define a Signed Local Similarity Indicator $\mc{I}\left(x;\mc{S}\right)$ as Definition~\ref{df:similarity}.
\begin{define}[Signed Local Similarity Indicator]\label{df:similarity}
Given a sample feature $x$, its Signed Local Similarity Indicator $\mc{I}\left(x;\mc{S}\right)$ is defined as 
\begin{align}\label{eq:indicator}
    \mc{I}\left(x;\mc{S}\right) & = \sum_{x_p\in\mc{S}^+}{\delta(m > w_{+}^{T}x_p - w_{-}^{T}x_p)x_p^Tx} \\\nonumber
    & - \sum_{x_n\in\mc{S}^-}{\delta(m > w_{-}^{T}x_n - w_{+}^{T}x_n)x_n^Tx}
\end{align}
where $\delta(\cdot)$ equals 1 if the condition inside holds otherwise equals 0.
\end{define}
From the definition, we see the indicator $\mc{I}(x;\mc{S})$ only focuses on the similarity between $x$ and those labeled samples $x_p, x_n$ close to classification boundary, i.e. samples that are vague for current classifiers to discriminate, when $x$ manifests stronger similarity with vague positive samples $x_p$ in batch $\mc{S}$, $\mc{I}(x;\mc{S})$ increases, in contrast, when $x$ is closer to vague negative samples, $\mc{I}(x;\mc{S})$ gets smaller value. With the Definition~\ref{df:similarity}, we claim that the Proposition~\ref{prop:mono} holds

\begin{prop}\label{prop:mono}\footnotemark[3]\footnotetext[3]{The proof can be found at supplementary material.}
If an unlabeled sample $x_t\in\mc{D}_t/\widetilde{\mc{D}}_t$ is measured by query function $Q(x_t)$ as Eq~(\ref{eq:query}), after a gradient descending step on batch $\mc{S}$, then the following monotonicity holds
\begin{itemize}
    \item if $p^+(x_t) > p^-(x_t)$, $Q(x_t)$ is decreasing monotonically with respect to $\mc{I}(x_t;\mc{S})$
    \item if $p^+(x_t) < p^-(x_t)$, $Q(x_t)$ is increasing monotonically with respect to $\mc{I}(x_t;\mc{S})$
\end{itemize}
\end{prop}

With the Proposition~\ref{prop:mono}, we see our margin loss performs under a mechanism analogous to Support Vector Machine~\cite{svm}, where only a few hard examples (like the support vectors) are collected as component to decide the query function score of a target sample $x_t$, e.g. if the trained classifier predicts that sample $x_t$ is more likely to be positive, i.e. $p^+(x_t) > p^-(x_t)$, then the closer $x_t$ is to existing hard positive samples, the less query function value $Q(x_t)$ will be obtained, in contrast, when $x_t$ is closer to some hard negative samples, margin loss will impose a larger $Q(x_t)$ score. 

\noindent\textbf{Transferability.} It should be noticeable that SDM is not only suitable for data selection, but also helps domain transfer theoretically. Following the analysis of~\cite{mme}, we define a margin-based domain classifier space as
\begin{equation}\label{eq:domain_cls}
    \mc{H}=\left\{ h(x)\right\} = \left\{\delta(|w_+^Tx - w_-^Tx| \geq m) | w_+,w_-\in \mc{R}^D \right\}
\end{equation}
then we can obtain the Proposition~\ref{prop:domain} to verify that SDM helps to shrink the domain gap~\cite{ben2010theory} under certain assumption
\begin{prop}\label{prop:domain}\footnotemark[3]
For source and target data $x_s \sim \mc{P}_s, x_t \sim \mc{P}_t$, given the margin domain classifier family $\mc{H}$ of Eq~(\ref{eq:domain_cls}), if $\mc{P}(h(x_t)=1) \leq \mc{P}(h(x_s)=1)$, then optimizing the binary margin loss is equivalent to minimize the upperbound of domain $\mc{H}$-divergency $d_{\mc{H}}(\mc{P}_s, \mc{P}_t)$ defined by~\cite{ben2010theory}. 
\end{prop}

\subsection{Variants}

\noindent\textbf{Dynamically Adjusted Margin Loss.} Although the loss in Eq~(\ref{eq:margin_loss}) can implicitly select hard source data, it still shows some flaws. First, all hard samples contributes equally in terms of the backward gradient. Besides, the margin constraint only considers the relative distance from samples to different class decision boundaries, ignoring constraint on the absolute score of ground-truth class label. With this consideration, we propose a dynamic version of margin loss to adaptively adjust the backward gradient in proportion to the margin size, and append a max-logit regularizer to ensure the gradient from ground-truth class will not vanish even if the margin is large enough than pre-defined $m$

\begin{align}\label{eq:margin_variant}
    \widetilde{\mc{L}}_m(x, y) & = \sum_{i\neq y}\alpha_i\left[m - c(g(x))_y + c(g(x))_i\right]_+ - c(g(x))_y \\
    \alpha_i & = 1-\frac{c(g(x))_y - c(g(x))_i}{m} \nonumber
\end{align}
In Eq~(\ref{eq:margin_variant}) we modify the rectified margin as a modulation factor $\alpha_i$ and take it to modulate the score of other classes except ground-truth. With this modulation, the loss term with smaller margin will be emphasized and generate larger gradient to push sample away from corresponding categorical clusters, helping our network adpatively focusing on hard source examples of different difficulties. Besides, a max-logit term is appended in Eq~(\ref{eq:margin_variant}) to constrain our network to always assign large score to prediction on ground-truth class. This kind of variant is termed as ``SDM-A''.

\begin{figure}
    \centering
    \includegraphics[width=0.46\textwidth]{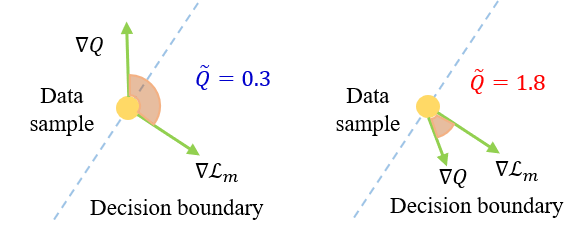}
    \caption{Examples of query with first-order differential margin. The left figure shows situation where the gradient direction from loss and query function diverge a lot. The right figure illustrates an example where the feature gradient from both loss and query function share similar update direction and yield high query score.}
    \label{fig:query}
\end{figure}
\noindent\textbf{Query with Gradient Direction Consistency.} To boost the selectivity of our query function, derived from the basic marge sampling in Eq~(\ref{eq:query}), we further take its variation into account. Inspired by~\cite{badge} which applies the weight variation to depict the data importance, we expect that the gradient from a newly appended sample will push its feature representation $\mb{f}$ toward direction that minimizes the margin sampling function as examplified in Figure~\ref{fig:query}, this is equivalent to ensure the gradient from both loss term and margin sampling manifests similar orientation in feature space
\begin{equation}\label{eq:query3}
    \widetilde{Q}(x) = Q(x) + \lambda\left<\nabla_{\mb{f}}{\mc{L}_m(x, y)}, \nabla_{\mb{f}}{Q_m(x)}\right>
\end{equation}
where $\left<\cdot,\cdot\right>$ is cosine-similarity metric, $\lambda$ is a balance factor. However, it is not possible to acquire the annotation $y$ of an unlabeled sample before selection, instead, we take the probabilistic gradient estimation $ \nabla_{\mb{f}}{\hat{\mc{L}}_m(x)}$ which is consistent with margin sampling
\begin{equation}
    \nabla_{\mb{f}}{\hat{\mc{L}}_m(x)} = \mb{p}_{1^*}\nabla_{\mb{f}}{\mc{L}_m(x, 1^*)} + \mb{p}_{2^*}\nabla_{\mb{f}}{\mc{L}_m(x, 2^*)}
\end{equation}
where the notation $\mb{p}, 1^*, 2^*$ follow the same definition from Eq~(\ref{eq:query}). Through the modified query function in Eq~(\ref{eq:query3}), the sampled data is not only close to decision boundaries of trained model, but also ensured to fast converged to a non-fuzzy state. This variant is termed as ``SDM-G''. Besides, the two variants are not mutually exclusive to each other and can be exploited simultaneously to obtain a combined active learning pipeline as ``SDM-AG''.

\begin{table*}
    \centering
    \resizebox{0.96\textwidth}{!}{$
        \begin{tabular}{c|cccccccccccc|c}
        \toprule
        \multirow{2}{*}{Method} & \multicolumn{12}{c}{Office-Home}\\
        & A $\to$ C & A $\to$ P & A $\to$ R & C $\to$ A & C $\to$ P & C $\to$ R & P $\to$ A & P $\to$ C & P $\to$ R & R $\to$ A & R $\to$ C & R $\to$ P & Avg\\
        \midrule
        ResNet~\cite{resnet} & 42.1 & 66.3 & 73.3 & 50.7 & 59.0 & 62.6 & 51.9 & 37.9 & 71.2 & 65.2 & 42.6 & 76.6 & 58.3\\
        \midrule
        RAN & 56.8 & 78.0 & 77.7 & 58.9 & 70.7 & 70.5 & 60.9 & 53.2 & 76.8 & 71.5 & 57.5 & 81.8 & 67.9\\
        ENT & 56.8 & 80.0 & 82.0 & 59.4 & 75.8 & 73.8 & 62.3 & 54.6 & 80.3 & 73.6 & 58.8 & 85.7 & 70.2\\
        CONF & 57.7 & 81.3 & 82.2 & 60.8 & 76.5 & 74.2 & 61.9 & 54.5 & 80.4 & 73.4 & 59.4 & 85.9 & 70.7\\
        MAR & 58.6 & 81.3 & 81.7 & 60.3 & 76.2 & 73.6 & 63.4 & 55.2 & 80.5 & 73.8 & 60.5 & 86.3 & 70.9\\
        QBC~\cite{qbc} & 56.9 & 78.0 & 78.4 & 58.5 & 73.3 & 69.6 & 60.2 & 53.3 & 76.1 & 70.3 & 57.1 & 83.1 & 67.9\\
        Cluster~\cite{cluster} & 56.0 & 76.8 & 78.1 & 58.4 & 72.6 & 69.2 & 58.4 & 51.2 & 75.4 & 70.1 & 56.4 & 82.4 & 67.1\\
        AADA~\cite{su2020active} & 56.6 & 78.1 & 79.0 & 58.5 & 73.7 & 71.0 & 60.1 & 53.1 & 77.0 & 70.6 & 57.0 & 84.5 & 68.3\\
        ADMA~\cite{adma} & 57.2 & 79.0 & 79.4 & 58.2 & 74.0 & 71.1 & 60.2 & 52.2 & 77.6 & 71.0 & 57.5 & 85.4 &68.6\\
        BADGE~\cite{badge} & 59.2 & 81.0 & 81.6 & 60.8 & 74.9 & 73.3 & 63.7 & 54.2 & 79.2 & 73.6 & 59.7 & 85.7 & 70.6\\
        TQS~\cite{fu2021transferable} & 58.6 & 81.1 & 81.5 & 61.1 & 76.1 & 73.3 & 61.2 & 54.7 & 79.7 & 73.4 & 58.9 & 86.1 & 70.5\\
        \midrule
        SDM-AG & \textbf{61.2} & \textbf{82.2} & \textbf{82.7} & \textbf{66.1} & \textbf{77.9} & \textbf{76.1} & \textbf{66.1} & \textbf{58.4} & \textbf{81.0} & \textbf{76.0} & \textbf{62.5} & \textbf{87.0} & \textbf{73.1}\\
        \bottomrule
        \end{tabular}
    $}
    \caption{Classification accuracy ($\%$) on the Office-Home dataset with the budget of 5\% data. Among the abbreviation, ``RAN'' is random sampling, ``ENT'' is entropy-based sampling, ``CONF'' is least confidence sampling and ``MAR'' is pure margin sampling.}
    \label{tab:officehome}
\end{table*}

\begin{table}
\small
    \centering
    \resizebox{0.47\textwidth}{!}{$
    \setlength{\tabcolsep}{1mm}{
    \begin{tabular}{c|cccccc|c}
    \toprule
    \multirow{2}{*}{Method} & \multicolumn{6}{c}{Office-31}\\
    & A$\to$W & A$\to$D & W$\to$A & W$\to$D & D$\to$A & D$\to$W & Avg\\
    \midrule
    ResNet~\cite{resnet} & 81.5 & 75.0 & 63.1 & 95.2 & 65.7 & 99.4 & 80.0\\
    \midrule
    RAN & 87.1 & 84.1 & 75.5 & 98.1 & 75.8 & 99.6 & 86.7\\
    UCN~\cite{unc} & 89.8 & 87.9 & 78.2 & 99.0 & 78.6 & 100.0 & 88.9\\
    QBC~\cite{qbc} & 89.7 & 87.3 & 77.1 & 98.6 & 78.1 & 99.6 &88.4\\
    Cluster~\cite{cluster} & 88.1 & 86.0 & 76.2 & 98.3 & 77.4 & 99.6 & 87.6\\
    AADA~\cite{su2020active} & 89.2 & 87.3 & 78.2 & 99.5 & 78.7 & 100.0 & 88.8\\
    ADMA~\cite{adma} & 90.0 & 88.3 & 79.2 & 100.0 & 79.1 & 100.0 & 89.4\\
    CLUE~\cite{CLUE} & 88.1 & 91.4 & 76.1 & 100.0 & 76.1 & 98.6 & 88.4\\
    TQS~\cite{fu2021transferable} & 92.2 & 92.8 & 80.4 & 100.0 & 80.6 & 100.0 & 91.0\\
    \midrule
    SDM-AG & \textbf{93.5} & \textbf{94.8} & \textbf{81.9} & \textbf{100.0} & \textbf{81.9} & \textbf{100.0} & \textbf{92.0}\\
    \bottomrule
    \end{tabular}}
    $}
    \caption{Classification accuracy ($\%$) on the Office-31 dataset with the budget of 5\% data. ``RAN'' represents random sampling.}
    \label{tab:office31}
    \vspace{-4mm}
\end{table}

\section{Experiments}
\subsection{Setup}
\noindent\textbf{Dataset and Metric.} In our experiments, We first evaluate the performance of our framework on two mainstream domain adaptation benchmarks,  Office-Home~\cite{saenko2010adapting} and Office-31~\cite{venkateswara2017deep}. Then we further extend our method to single-domain dataset CIFAR-10~\cite{krizhevsky2009learning} to validate the generality of SDM. The Office-31 dataset includes 3 different domain with imbalance image distribution, there are total 4110 images of 31 object categories. The Office-Home dataset is a more challenging benchmark consisting of 4 different domains and 65 different types of objects. CIFAR-10 is a widely used dataset for different machine learning tasks, there are total 50000 images for 10 common classes. Following the work of~\cite{fu2021transferable}, for experiments on Office-31 and Office-Home, we report results on all transfer scenarios and average the accuracy on different scenarios for final comparison. Our active learning loop starts with data from only source domain, at each sampling step, $1\%$ of target data is sampled, and totally $5$ times of sampling steps are conducted. For CIFAR-10, our training process starts from $10\%$ of full training data, at each sampling step, $5\%$ of data is sampled and the budget is set as $30\%$ of training data.

\noindent\textbf{Implementation Detail.}
Our experiments are implemented with Pytorch framework. Following the setting of~\cite{fu2021transferable}, we take the commonly used ResNet50~\cite{resnet} architecture which is pre-trained on ImageNet~\cite{krizhevsky2012imagenet} as our feature extractor and classifier. Different from some of previous methods for ADA~\cite{fu2021transferable,CLUE} combining unsupervised domain adaption methods and trained with data from target domain, in our implementation, we avoid training on unlabeled data with any unsupervised learning technique for fairer comparison, this also makes our SDM suitable for both pooled and sequential setting of active learning. During the training process, we first train our network with initial data for 10 epochs with margin loss and an auxiliary cross entropy loss, after which we start our sampling steps. The sampling process is performed every two epochs until the labeled target data reaches the total budget. The learning rate is set to be $0.01$ and batch size is set as $72$. We set the hyper parameter margin $m$ in Eq~(\ref{eq:margin_loss}) to 1 and $\lambda$ in Eq~(\ref{eq:query3}) to $0.01$ in terms of detailed ablation studies. 

\begin{figure}
    \centering
    \includegraphics[width=0.48\textwidth]{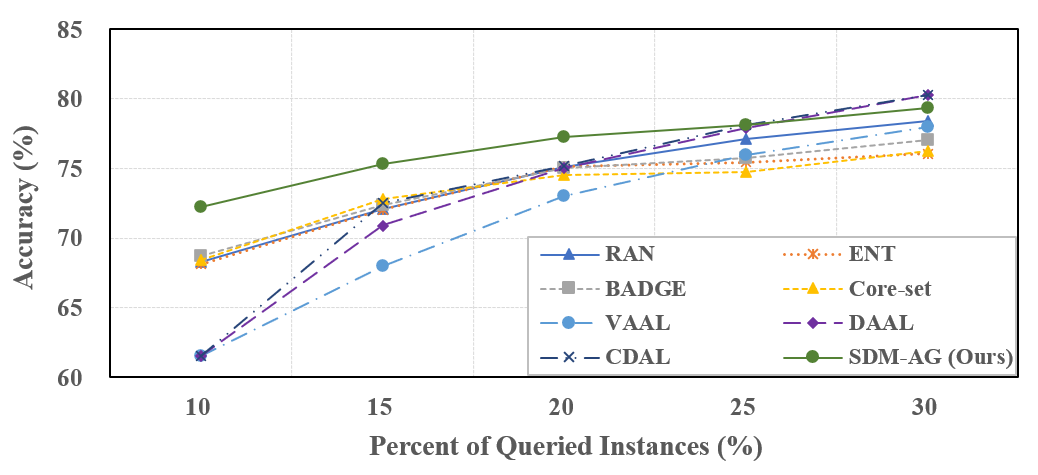}
    \caption{Experiment results on CIFAR-10 dataset from 10\% training data to 30\% training data. ``RAN'' is random sampling and ``ENT'' is entropy-based sampling.}
    \label{fig:cifar}
    \vspace{-5mm}
\end{figure}

\subsection{Main Results}

\begin{table*}
    \centering
    \resizebox{\textwidth}{!}{
        \begin{tabular}{c|cc|cccccccccccc|c}
        \toprule
        Method & Adjust & Gradient & A$\to$C & A$\to$P & A$\to$R & C$\to$A & C$\to$P & C$\to$R & P$\to$A & P$\to$C & P$\to$R & R$\to$A & R$\to$C & R$\to$P & Avg\\
        \midrule
        Baseline & & & 58.6 & 81.3 & 81.7 & 60.3 & 76.2 & 73.6 & 63.4 & 55.2 & 80.5 & 73.8 & 60.5 & 86.3 & 70.9\\
        SDM & & & 60.5 & 79.6 & 81.4 & 65.3 & 76.5 & 74.9 & 65.8 & 56.5 & 80.6 & 75.2 & 61.1 & 85.7 & 71.9 \\
        \midrule
       SDM-A  & $\checkmark$ & & 60.7 & 81.5 & 82.1 & 65.7 & 76.8 & \textbf{76.3} & \textbf{66.3} & 58.1 & 80.2 & 75.2 & \textbf{62.7} & 86.6 & 72.7\\
       SDM-G & & $\checkmark$ & \textbf{61.2} & 81.9 & \textbf{82.7} & 65.6 & 77.6 & 76.1 & 66.0 & 58.0 & 80.8 & 75.8 & 61.8 & 86.9 & 72.9\\
       SDM-AG & $\checkmark$ & $\checkmark$ & \textbf{61.2} & \textbf{82.2} & \textbf{82.7} & \textbf{66.1} & \textbf{77.9} & 76.1 & 66.1 & \textbf{58.4} & \textbf{81.0} & \textbf{76.0} & 62.5 & \textbf{87.0} & \textbf{73.1}\\
        \bottomrule
        \end{tabular}
    }
    \caption{Ablation study with different configuration with $5\%$ of target labeled data on Office-Home dataset. ``Baseline'' is a model trained with cross-entropy loss and selecting data with margin sampling. ``Adjust'' indicates dynamically adjusted margin loss. ``Gradient'' denotes first-order gradient consistency.}
    \label{tab:ablation_study}
\end{table*}

\begin{table}
    \centering
    \begin{tabular}{c|c|c|c}
    \toprule
    Sample Strategy & Training Loss & Acc & $\Delta$ \\
    \midrule
    \multirow{2}{*}{Entropy} & Cross Entropy & 70.24 & \multirow{2}{*}{+0.28}\\
    & Margin Loss & \textbf{70.52}\\
    \midrule
    \multirow{2}{*}{Least Confidence} & Cross Entropy & 70.68 & \multirow{2}{*}{+0.54}\\
    & Margin Loss & \textbf{71.22}\\
    \midrule
    \multirow{2}{*}{Margin Sample} & Cross Entropy & 70.94 & \multirow{2}{*}{\textbf{+0.98}}\\
    & Margin Loss & \textbf{71.92}\\
    \bottomrule
    \end{tabular}
    \caption{Comparison with different combination between different types of training loss and sampling strategies on Office-Home dataset. The ``Acc'' represents the averaged accuracy over all 12 transferring scenarios. $\Delta$ denotes improvement with margin loss.}
    \label{tab:compare_with_ce}
\end{table}

We compare our ``SDM-AG'' pipeline with other active learning approaches on different benchmarks. We take a ResNet50 trained with pure initial source data as our baseline method for comparison, methods with classical active learning strategies~\cite{qbc,unc,cluster,badge} are taken into account, further, we also compare our methods with recent state-of-the-art ADA approaches~\cite{fu2021transferable,adma,su2020active,CLUE}. Besides, we also compare with some commonly used simple query functions like random sampling (RAN), entropy-based sampling (ENT), least confidence (CONF) and margin sampling (MAR). 

The comparison results on Office-Home are presented in Table~\ref{tab:officehome}. From this table, we see our SDM-AG pipeline outperforms either classical active learning approaches or recent ADA methods designed with complicated selection strategies. To be specific, our SDM-AG method can bring about $+2.6\%$ performance gain in average accuracy over state-of-the-art active learning methods like TQS~\cite{fu2021transferable} or BADGE~\cite{badge}. Further, it can be observed that in some harder scenarios with larger discrepancy between source and target (e.g. \emph{C to A} and \emph{P to A}), the improvement from our SDM-AG method is more salient. In total, our method can achieve $+14.8\%$ improvement on the average performance over the baseline with pure source data. Similar results can be found on the dataset of Office-31, which is listed in Table~\ref{tab:office31}. Although some transferring scenarios in this benchmark is kind of saturated, it can still be observed that SDM-AG achieves substantial performance gain over other state-of-the-art method~\cite{CLUE,su2020active,fu2021transferable} on some challenging scenarios, and our simple pipeline can outperforms all compared methods in terms of averaged domain adaptation accuracy.

In addition, we also extend our experiment and comparison to a general active learning setting without domain gap on the benchmark of CIFAR-10. The results are evaluated after training of each sampling step and ploted in Figure~\ref{fig:cifar}. It can be observed that our SDM pipeline can still outperforms most of other state-of-the-art methods~\cite{sener2018active,badge,sinha2019variational} regardless of numbers of labeled data and comparable to some newest AL algorithms~\cite{wang2020dual,agarwal2020contextual}. It is also noticeable that when the number of queried number is small (e.g. $10\%\sim 20\%$ of training data), SDM outperforms all competitors including recently proposed DAAL~\cite{wang2020dual} or CDAL~\cite{agarwal2020contextual} by a large margin, demonstrating our SDM algorithm is more friendly to scenarios of active learning with low budget.

\subsection{Detailed Analysis}

In this section, we analysis the components of our algorithm in detail. If not specified, the analysis is conducted on Office-Home with our default setting.

\noindent\textbf{Improvement over Cross-Entropy Baseline.} First we conduct experiments to investigate the superiority of SDM over a simple active learning baseline. To thie end, we design a baseline method where the network is trained with pure cross-entropy loss, but selecting samples with the same criterion of margin sampling as Eq~(\ref{eq:query}). The comparison on different transferring scenarios is shown in Table~\ref{tab:ablation_study}. It is observed that our SDM paradigm, i.e. model trained with margin loss and selecting data by margin sampling achieves consistent improvement over cross-entropy baseline on most scenarios and overall performance. This observation indicates that the improvement is the results of the whole solution of SDM instead of a simple inclusion of sampling strategy.

\noindent\textbf{Effectiveness of Different Variants.} Next we investigate the improvement of different variants based on our SDM baseline. The results are listed in Table~\ref{tab:ablation_study}. From the table, we observe that both SDM-A and SDM-G can bring significant performance gain compared with simple SDM pipeline, demonstrating the improved dynamic margin loss and query function with gradient guidance can benifit the active learning process respectively. Besides, we see the combination of two variants, i.e. SDM-AG can further boost the averaged performance to at most $73.1\%$ and achieve the best results on most of scenarios of Office-Home dataset.

\noindent\textbf{Compatibility between Margin Loss and Sampling.} In the discussion of Proposition~\ref{prop:mono}, we have show training with margin loss is inherently helpful for margin sampling especially to mine informative data from target domain. In this section we further investigate this property with empirical results. To this end, we test different combination between training objective and query functions. For training loss, we investigate the margin loss and commonly used cross-entropy loss, as for sampling strategy, in addition to margin sampling, the commonly used least confidence and entropy sampling strategies are exploited. The test results are shown in Table~\ref{tab:compare_with_ce}, from the table we can conclude that: (1) Regardless of the sampling strategy we use, margin loss can bring improvement over pipeline trained with cross-entropy. (2) In terms of the performance gain ($\Delta$ in Table~\ref{tab:compare_with_ce}), the margin sampling strategy obtains the most gain in average accuracy, indicating that the margin loss is inherently suitable for a relative margin-based data selection strategy to mine informative data for domain transfer, which is consistent to Proposition~\ref{prop:mono}.

\begin{figure}
\centering
    \hspace{-3mm}
    \subfloat[P to C]{
    \label{fig:budget_pc}
    \includegraphics[width=0.233\textwidth]{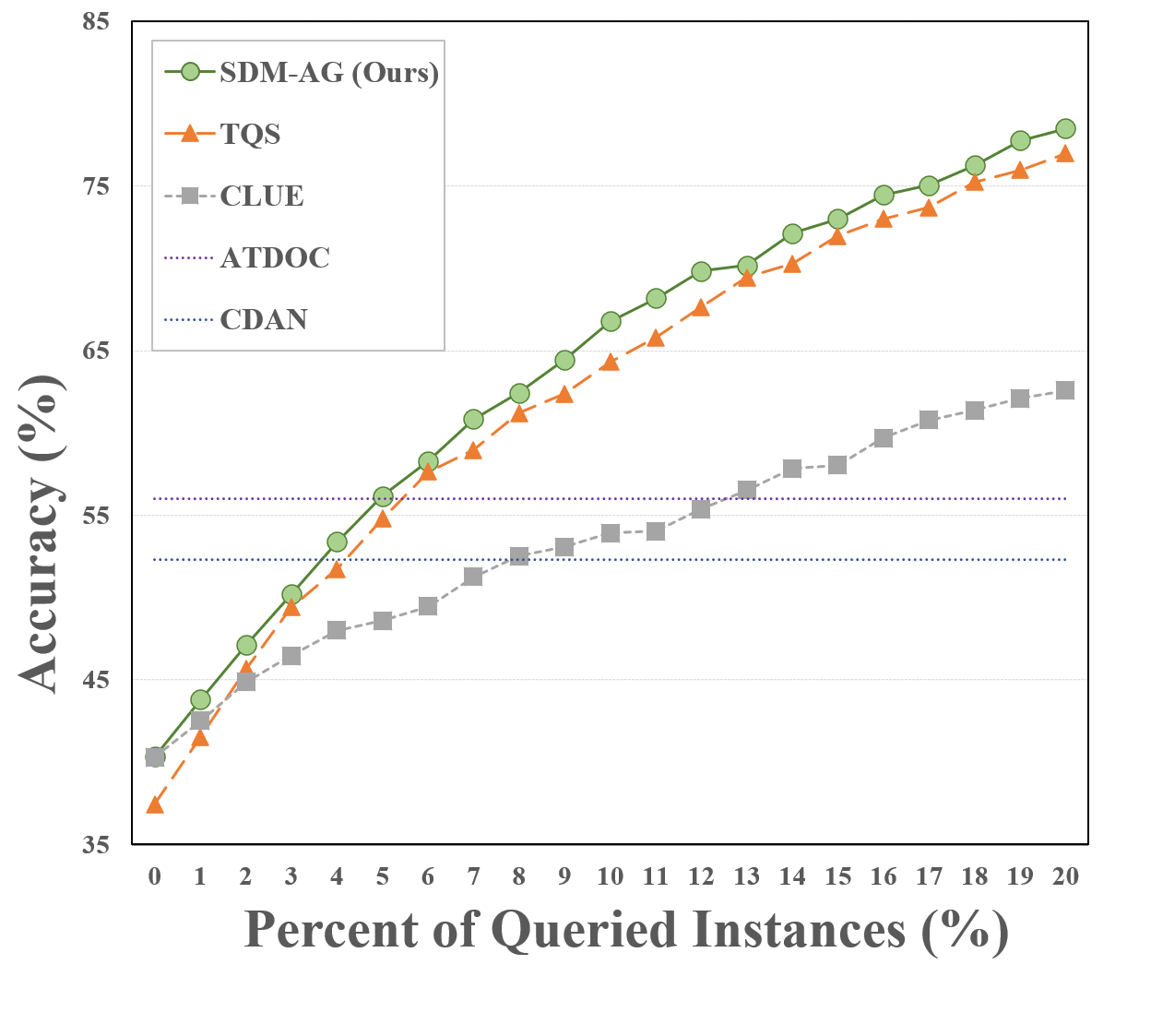}
    }
    \hspace{-3mm}
    \subfloat[R to A]{
    \label{fig:budget_ra}
    \includegraphics[width=0.233\textwidth]{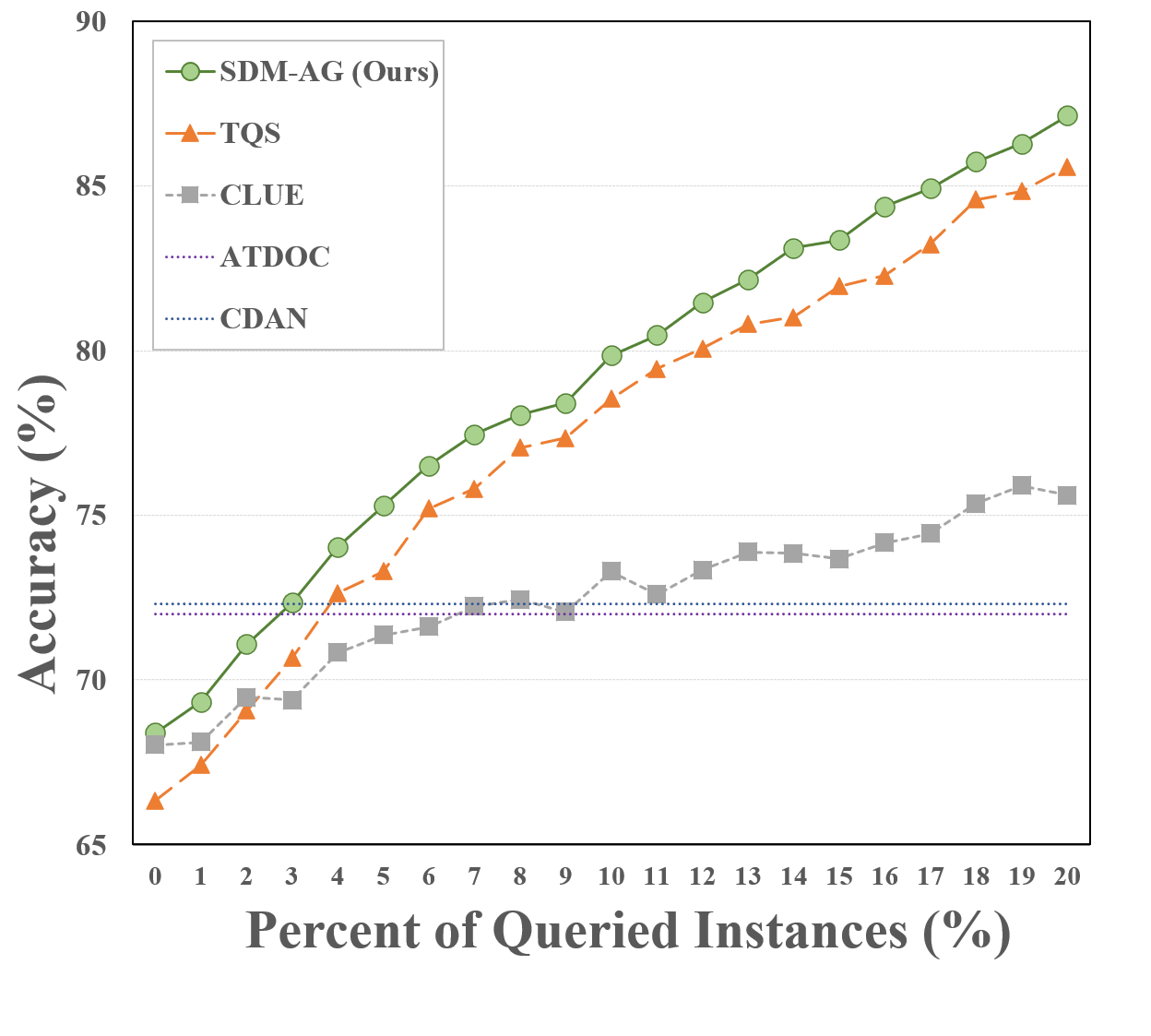}
    }
    \caption{Performance variation with different budget size on different scnarios of Office-Home dataset.}
    \label{fig:budget}
\end{figure}
\begin{figure}
\centering
    \hspace{-3mm}
    \subfloat[Analysis of $m$]{
    \label{fig:sense-m}
    \includegraphics[width=0.233\textwidth]{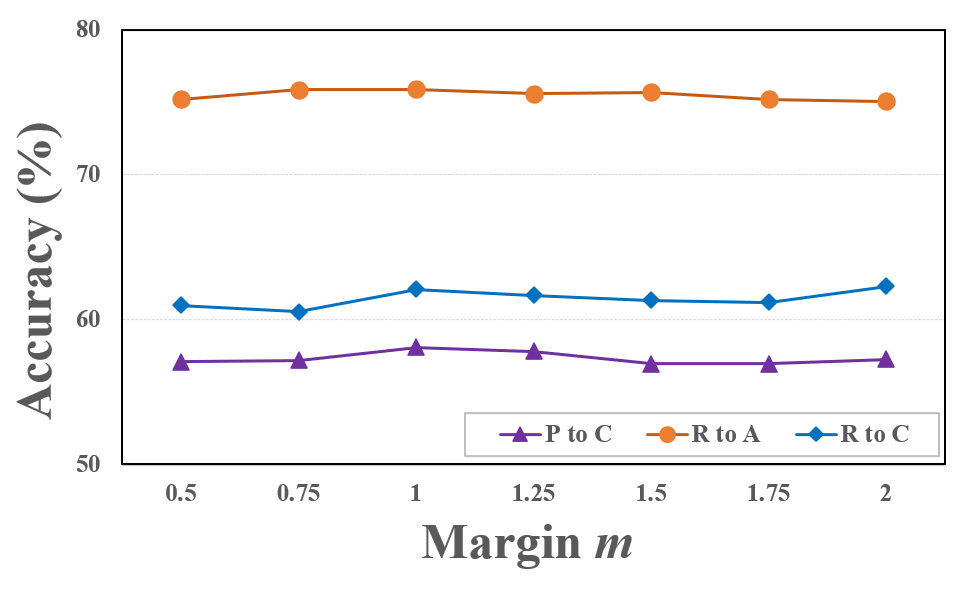}
    }
    \hspace{-3mm}
    \subfloat[Analysis of $\lambda$]{
    \label{fig:sense-lambda}
    \includegraphics[width=0.233\textwidth]{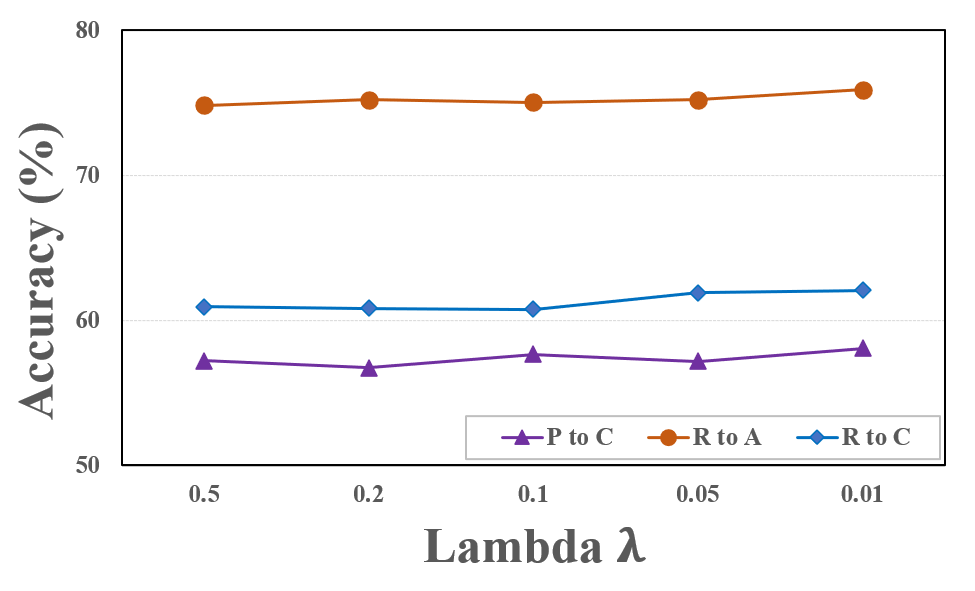}
    }
    \caption{Sensitivity analysis on hyper-parameters of SDM on different scenarios of Office-Home dataset.}
    \label{fig:sensitivity}
    \vspace{-5mm}
\end{figure}

\noindent\textbf{Variation with Different Budget Size.} The annotation budget $B$ is an important parameter for active learning since it decides the available target data to be labeled, therefore we test how the domain adaptation performance varies with increasing budget. To make horizontal comparison, we also compare SDM with other two ADA methods TQS~\cite{fu2021transferable} and CLUE~\cite{CLUE}. In addition, we also compare with recent unsupervised DA methods like CDAN~\cite{cdan} and ATDOC~\cite{atdoc} to see the required sample number to achieve competitive results to these approaches. The budget size $B$ is controlled within the range from $0\%$ to $20\%$ of target data. The results are plotted as curves in Figure~\ref{fig:budget}. When compared with TQS and CLUE, our SDM-AG methods can achieve consistent improvement regardless of the budget size, this superiority is more obvious on the scenario of \emph{R to A}, demonstrating that our method can steadily benefit from the growth of budget $B$ and is not easy to saturate. Besides, when compared with ATDOC and CDAN, our methods can achieve the comparable results with a burden of only $5\%$ target data labeled, demonstrating the efficiency of our algorithm.

\noindent\textbf{Sensitivity of Hyper-parameter.} We further test how the hyper-parameters in our SDM pipeline affect the overall performance of domain adaptation to see if our algorithm is sensitive to some parameters. To be specific, we tune the training margin $m$ of Eq~(\ref{eq:margin_variant}) and balance factor $\lambda$ in Eq~(\ref{eq:query3}) within a tolerable range and test the accuracy on three scenarios of different difficulties (\emph{P to C}, \emph{R to C}, \emph{R to A}). The results are ploted as curves in Figure~\ref{fig:sensitivity}. On all scenarios, we see the accuracy varies marginally with the tuned hyper-parameters. This observation demonstrate our approach is stable and not sensitive to specific hyper-parameters.

\noindent\textbf{Complexity Analysis.} Finally, we analyze the complexity and running time to confirm the claim that our SDM algorithm is a simple pipeline compared with other complicated ADA methods. To be specific, we compare the theoretical complexity and actual running time of one round of data query and sampling. We compare SDM with state-of-the-art clustering method~\cite{CLUE,badge} and ranking method~\cite{fu2021transferable}. For all methods, we ignore the running time and complexity of network forward pass since this is the common step and consumes the same time, and for all rank-based methods, we assume a stable comparison sort algorithm is applied with the lower bound of complexity $\mc{O}(N\log N)$ to sort all data. The comparison results are listed in Table~\ref{tab:complexity}, readers can refer to the appendix material for more details about the complexity derivation of SDM. In Table~\ref{tab:complexity}, we see the rank-based methods do not rely on budget size $B$, resulting more efficient complexity. In terms of running time, SDM achieves $24.6\times$ query speed compared with the nearest competitor~\cite{CLUE} and is much faster than TQS~\cite{fu2021transferable}, since TQS requires parsing results from multiple classifiers and running an additional discrimination network for domainess.

\begin{table}
    \centering
    \begin{tabular}{c|c|c}
    \toprule
     Method & Query Complexity & Time (s) \\
    \midrule
    BADGE~\cite{badge} & $\mc{O}(BNKD)$ & 11.47 \\
    CLUE~\cite{CLUE} & $\mc{O}(tNBD)$ & 1.65 \\
    \midrule
    TQS~\cite{fu2021transferable} & $\mc{O}(NMK+N\log N)$ & 2.19 \\
    SDM-AG (ours) & $\mc{O}(NKD+N\log N)$ & \textbf{0.067}\\
    \bottomrule
    \end{tabular}
    \caption{Comparison between complexity and running time of different methods. $B$ is the budget size, $K$ is number of classes, $D$ is feature dimension, $N$ is number of target samples, $t$ is the clustering iteration in~\cite{CLUE} and $M$ is the committee size in~\cite{fu2021transferable}.}
    \label{tab:complexity}
    \vspace{-5mm}
\end{table}

\section{Conclusion}

In this paper, aimed at the active domain adaptation problem, we propose a simple but effective solution termed as Select-by-Distinctive-Margin (SDM). We provide theoretical analysis to show how a model trained with margin loss select informative data, and further propose two variants to enhance the model training and data sampling. Comprehensive experiment results demonstrate that our algorithm is a concise, stable and superior solution toward the active domain adaptation problem.

\section*{Acknowledgement}
This work is supported by following foundations: National Natural Science Foundation of China No. 11988101, Natural Science Foundation of China under contract 62171139, National Key R\&D Program under contract (2017YFA0402600), National Natural Science Foundation of China under grant U2031117, the Youth Innovation Promotion Association CAS (id. 2021055), CAS Project for Young Scientists in Basic Reasearch (grant YSBR-006) and the Cultivation Project for FAST Scientific Payoff and Research Achievement of CAMS-CAS.


{\small
\bibliographystyle{ieee_fullname}
\bibliography{egbib}
}

\end{document}